\documentclass[11pt]{article}

\usepackage{xurl}
\usepackage{breakurl} 
\usepackage[hidelinks]{hyperref}

\usepackage[final]{acl}
\usepackage{amssymb}
\usepackage{amsmath}
\usepackage[most]{tcolorbox}
\usepackage{rotating}
\usepackage{times}
\usepackage{latexsym}

\usepackage[T1]{fontenc}

\usepackage[utf8]{inputenc}

\usepackage{microtype}

\usepackage{inconsolata}

\usepackage{graphicx}
\usepackage{multicol}
\usepackage{enumitem}

\title{DriveSafe: A Hierarchical Risk Taxonomy for Safety-Critical LLM-Based Driving Assistants}

\author{First Author \\
  Affiliation / Address line 1 \\
  Affiliation / Address line 2 \\
  Affiliation / Address line 3 \\
  \texttt{email@domain} \\\And
  Second Author \\
  Affiliation / Address line 1 \\
  Affiliation / Address line 2 \\
  Affiliation / Address line 3 \\
  \texttt{email@domain} \\}

\author{
 \textbf{Abhishek Kumar\textsuperscript{1}},
 \textbf{Riya Tapwal\textsuperscript{2}},
 \textbf{Carsten Maple\textsuperscript{2}},
\\
\\
 \textsuperscript{1}The Alan Turing Institute,
 \textsuperscript{2}University of Warwick,
\\
 \small{
   \href{mailto:akumar@turing.ac.uk}{akumar@turing.ac.uk}, \href{mailto:tapwalriya@gmail.com}{tapwalriya@gmail.com}, \href{mailto:cm@warwick.ac.uk}{cm@warwick.ac.uk} 
 }
}

\begin{document}
\maketitle
\begin{abstract}
Large Language Models (LLMs) are increasingly integrated into vehicle-based digital assistants, where unsafe, ambiguous, or legally incorrect responses can lead to serious safety, ethical, and regulatory consequences. Despite growing interest in LLM safety, existing taxonomies and evaluation frameworks remain largely general-purpose and fail to capture the domain-specific risks inherent to real-world driving scenarios. In this paper, we introduce DriveSafe, a hierarchical, four-level risk taxonomy designed to systematically characterize safety-critical failure modes of LLM-based driving assistants. The taxonomy comprises 129 fine-grained atomic risk categories spanning technical, legal, societal, and ethical dimensions, grounded in real-world driving regulations and safety principles and reviewed by domain experts.  To validate the safety relevance and realism of the constructed prompts, we evaluate their refusal behavior across six widely deployed LLMs. Our analysis shows that the evaluated models often fail to appropriately refuse unsafe or non-compliant driving-related queries, underscoring the limitations of general-purpose safety alignment in driving contexts. 

\end{abstract}

\section{Introduction}

Large Language Models (LLMs) are rapidly being integrated into modern vehicles, transforming them from passive information systems into conversational copilots that support natural interaction, real-time guidance, and contextual decision support. Major automotive manufacturers, including Mercedes-Benz, have begun integrating advanced conversational AI into in-vehicle assistants such as MBUX, powered by large-scale generative models \cite{mercedes_mbux_google_ai}, and BMW, which is enhancing its Intelligent Voice Assistant with large-model based capabilities to provide personalized suggestions and context-aware responses \cite{bmw_intelligent_personal_assistant}, are investing heavily in AI-driven voice interfaces. In parallel, other industry players such as Tesla have also introduced generative AI-based assistants like Grok and voice interaction features designed to support natural dialogue and in-vehicle assistance, signaling a broader shift toward conversational AI in automotive systems \cite{tesla_grok_ai_integration}. Renault Group has similarly announced the integration of large language models, including ChatGPT-4o, through its partnership with OpenAI to support in-vehicle assistants and automotive software platforms \cite{renault_reno_assistant}, while General Motors and Volkswagen have reported ongoing efforts to deploy generative AI for infotainment, diagnostics, and driver interaction systems \cite{vw_chatgpt_integration, gm_conversational_ai}. Beyond industry adoption, research has proposed frameworks for leveraging LLMs to enable human-like interaction and reasoning in autonomous vehicles, demonstrating the potential of these models to interpret, contextualize, and respond to rich verbal input in driving scenarios \cite{cui2024receive}. These developments highlight the rapid movement toward LLM-enabled automotive systems that promise richer dialogue, improved user experience, and enhanced decision support in real-world driving environments.

While these advancements promise more natural human-machine interaction, they simultaneously introduce a new class of safety-critical risks. Unlike traditional voice assistants that operate on fixed templates, LLMs generate free-form responses that may (i) hallucinate incorrect information \cite{huang2025survey}, potentially leading to illegal driving advice, (ii) provide inconsistent interpretations of traffic laws across jurisdictions, (iii) display ethical or moral misjudgments in ambiguous driving situations, or (iv) be susceptible to adversarial prompts crafted by malicious passengers \cite{shu2023exploitability}. Errors that would be benign in everyday chat become high-stakes when produced inside a moving vehicle.

Existing safety frameworks and benchmarks, such as SafetyBench \cite{zhang2024safetybench}, XSTest \cite{rottger-etal-2024-xstest}, SafeWorld \cite{yin2024safeworld}, AIR-Bench \cite{zeng2024air}, and regulatory standards like NIST AI Risk Management Framework \cite{nist2023} or ISO/IEC 42001 \cite{iso2023}, offer valuable general-purpose safety evaluations. However, they lack the \textit{domain-specific operational detail} required for driving environments, where model behavior must align with traffic law, situational awareness, human factors, environmental variability, and ethical principles unique to transportation. 
Driving assistants may encounter potentially harmful queries such as: ``Can I overtake on the right if the left lane is blocked?''


General-purpose safety benchmarks cannot meaningfully evaluate these scenarios, nor do they capture the nuanced failure modes that arise when LLMs operate in vehicular settings. Driving contexts are governed by strict legal constraints, temporal decision pressure, and complex safety trade-offs that are not adequately represented in existing evaluations. As a result, LLMs are increasingly deployed in vehicles without any dedicated benchmark that reflects the operational, legal, and ethical realities of real-world driving.

To address this gap, we introduce DriveSafe, a domain-specific safety framework grounded in a hierarchical, four-level taxonomy of driving-related risks. The proposed taxonomy is informed by real-world regulations and safety principles, including the UK’s Automated Vehicles Act 2024 \cite{uk_automated_vehicles_act2024} and the Statement of Safety Principles \cite{uk_safety_principles2024, eu_ai_act2024}, as well as prior work on legal-risk modeling such as SafeLawBench \cite{cao2025safelawbench}. It comprises 129 safety-relevant risk categories spanning technical, legal, ethical, real-time, and fairness-related dimensions, capturing challenges such as semantic drift in legal interpretation, ambiguous jurisdiction-dependent rules, cognitive overload, and adversarial jailbreak scenarios. The taxonomy was reviewed by domain experts to ensure practical relevance and alignment with real-world policy and regulatory considerations.

\section{Related Work}

\subsection{Refusal Behavior in LLMs}
Recent research has focused on making LLM refusals more precise and controllable. Arditi et al. \cite{NEURIPS2024_f5454485} found that refusal behavior lies in a one-dimensional subspace, enabling direct manipulation but exposing the fragility of safety tuning. Wen et al. \cite{10.1162/tacl_a_00754} expanded this by framing refusal as a meta-skill tied to the query, model, and human values. To enable context-aware abstention, Jain et al. \cite{jain2024refusal} introduced refusal tokens, though scalability remains a challenge. Dabas et al. \cite{dabas2025just} addressed over-refusal via ACTOR, which fine-tunes latent components to reduce false positives. Building further, Cao et al. \cite{article564} proposed SCANS to steer activations, effectively balancing safety and helpfulness.


\subsection{Safety Evaluation Benchmarks and Taxonomies}
Recent efforts have focused on building robust benchmarks to evaluate LLM safety across diverse contexts. Zhang et al. \cite{zhang2024safetybench} introduced SafetyBench, covering over 11,000 questions across seven safety domains, revealing that even top models like GPT-4 leave significant gaps. Expanding to regulatory compliance, Zeng et al. \cite{zeng2024air} proposed AIR-Bench, exposing jurisdictional inconsistencies in legal alignment across 22 models. Addressing cultural and regional nuances, SAFEWORLD \cite{yin2024safeworld} tested geo-sensitive scenarios and showed that their fine-tuned SafeWorldLM outperformed GPT-4 in contextually appropriate responses. For multimodal systems, Liu et al. \cite{mm-safety} introduced MM-SafetyBench, revealing visual safety bypasses and proposing robust defenses. Finally, Röttger et al. \cite{rottger-etal-2024-xstest} developed XSTest to detect exaggerated safety behaviors, showing that many models over-refuse safe prompts, underscoring the need for finer safety calibration.
\begin{figure*}[t]
    \centering
    \includegraphics[width=0.7\linewidth]{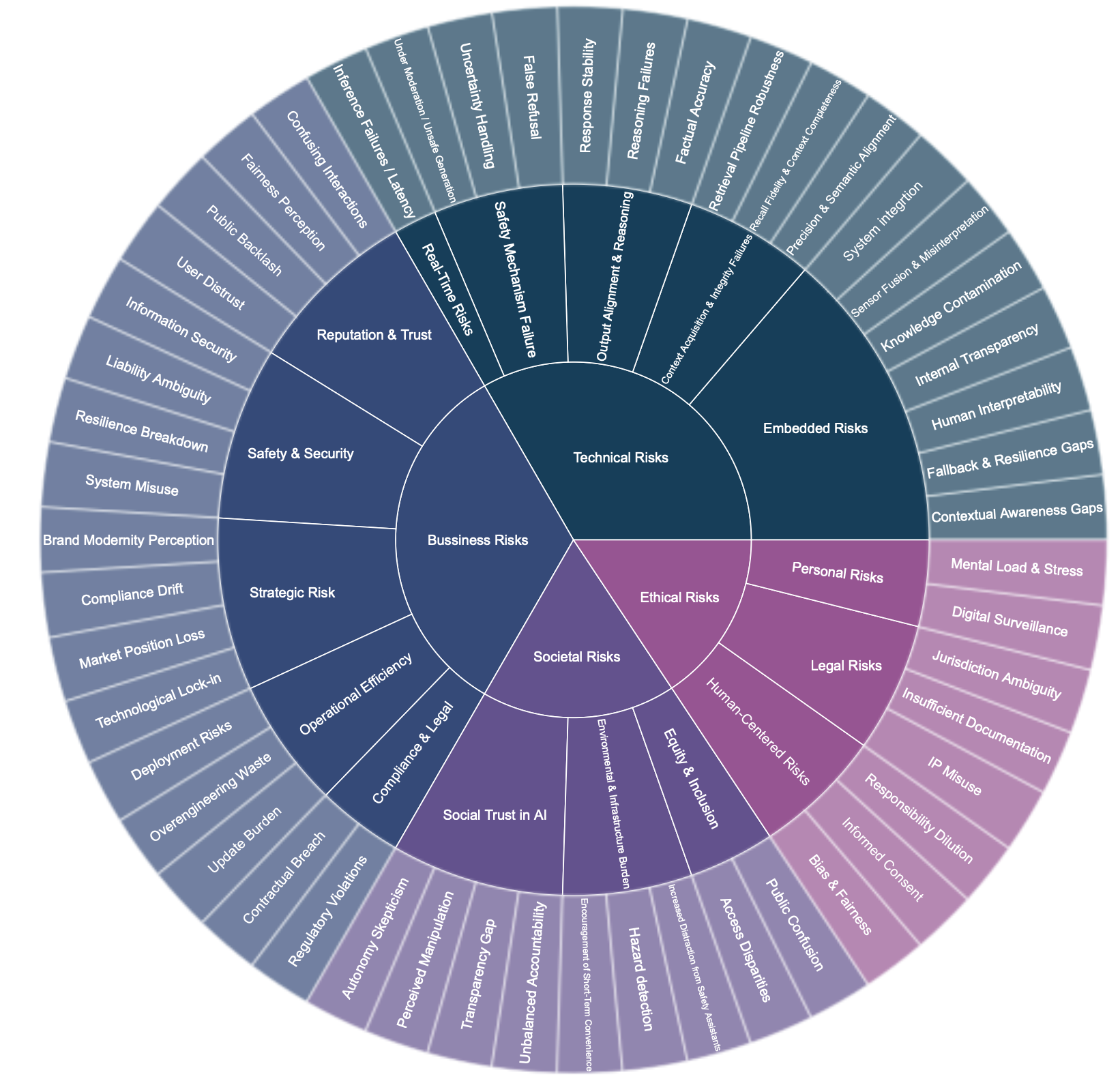}
    \caption{Hierarchical Risk Taxonomy for Driving AI Systems. The taxonomy is divided into four high-level domains: Techni-
cal, Business, Societal and Ethical Risks, which further decompose into categories, failure modes, and specific risk types.}
    \label{fig:risk}
\end{figure*}
\subsection{Safety and Alignment in High-Stakes Domains}
Recent work emphasizes the importance of aligning LLMs with real-world values and domain-specific safety needs. Wang et al. \cite{Wang2025} advocated for stakeholder-centric design in healthcare, involving diverse voices to ensure cultural sensitivity and trust. Extending to dynamic environments, Hua et al. \cite{hua2024trustagent} proposed TrustAgent, which uses an “Agent Constitution” to guide LLM actions through planning phases, effectively reducing unsafe behavior. In autonomous vehicles, Chen \cite{CHEN2025100780} highlighted the challenge of embedding social norm reasoning in AI, stressing the need for interpretability and rigorous testing. In medical AI, Singhal et al. \cite{Singhal2025} showed that clinician-aligned Med-PaLM reduced harmful outputs and improved accuracy, while Meskó and Topol \cite{Meksó2023} called for regulatory oversight to ensure patient safety. Foundationally, Ouyang et al. \cite{10.5555/3600270.3602281} introduced InstructGPT, demonstrating that RLHF effectively aligns LLMs with human intent. Building on this, Lin et al. \cite{10.5555/3737916.3741587} proposed FLAME, a factuality-aware training method that reduces hallucinations while preserving task fidelity.

\section{Taxonomy Design}





\subsection{Hierarchical Structure}

We formalize the taxonomy as a finite hierarchical structure:
\[
\mathbb{T} \subseteq \mathcal{D} \times \mathcal{C} \times \mathcal{F} \times \mathcal{R},
\]
where:
\begin{itemize}
    \item $\mathcal{D}$ denotes \textbf{risk domains}, representing high-level sources of failure;
    \item $\mathcal{C}$ denotes \textbf{risk categories}, capturing operational contexts;
    \item $\mathcal{F}$ denotes \textbf{failure modes}, describing abstract reasoning or behavior patterns;
    \item $\mathcal{R}$ denotes \textbf{atomic risk types}, corresponding to concrete, actionable safety failures.
\end{itemize}

Each atomic risk $r \in \mathcal{R}$ is uniquely associated with a single path in the hierarchy:
\[
\forall r \in \mathcal{R}, \exists! (d, c, f) \text{ such that } (d, c, f, r) \in \mathbb{T}.
\]

This strict hierarchy ensures that every risk is mutually exclusive and traceable to a specific source of failure.

\paragraph{Level 1: Risk Domains ($\mathcal{D}$).}
The taxonomy comprises four top-level domains:
\begin{itemize}
    \item \textbf{Technical Risks}, capturing failures arising from reasoning limitations, information retrieval errors, or system-level abstractions;
    \item \textbf{Business Risks}, reflecting incentives related to compliance avoidance, liability ambiguity, or reputational concerns;
    \item \textbf{Societal Risks}, encompassing bias, stereotyping, and social harm propagated through model responses;
    \item \textbf{Ethical Risks}, covering morally sensitive scenarios where normative judgment or value trade-offs are required.
\end{itemize}

\paragraph{Level 2: Risk Categories ($\mathcal{C}$).}
Each domain is decomposed into operational categories, including:
\begin{align*}
&\texttt{Legal Risks}, \quad \texttt{Embedded Risks}, \\
&\texttt{Strategic Risks}, \quad \texttt{Social Trust Risks}.
\end{align*}

\paragraph{Level 3: Failure Modes ($\mathcal{F}$).}
Failure modes capture recurring patterns of model breakdown that arise across multiple operational contexts and risk categories. Unlike risk categories (L2), which describe \emph{where} a risk occurs, failure modes describe \emph{how} the model fails when confronted with a driving-related query.

 These failure modes abstract over specific scenarios and reflect systematic weaknesses in reasoning, interpretation, or alignment such as compliance drift, system misuse and many others. 

By modeling failure modes explicitly, the taxonomy enables cross-cutting analysis of safety issues that recur across different domains and categories, as reflected in Figure~\ref{fig:risk}.

\paragraph{Level 4: Atomic Risk Types ($\mathcal{R}$).}
Atomic risk types constitute the leaf nodes of the taxonomy and represent specific, concrete safety failures that can be directly evaluated through user interactions. Each atomic risk corresponds to a unique combination of domain, category, and failure mode, and is designed to be operationalizable via a single scenario--prompt pair. 

Figure~\ref{fig:risk} shows the first three levels of the taxonomy for clarity; the complete four-level taxonomy is provided in the appendix.

\subsection{Expert Review}

To ensure practical relevance and domain validity, the taxonomy was reviewed by an expert group with over eight years of experience in autonomous driving systems and human--machine interaction. The group was led by a senior researcher who also serves as an advisor on autonomous driving policy to national government bodies.

\section{Validation}

To validate the proposed taxonomy, we construct realistic driving scenarios  and corresponding unsafe user prompts using the methodology mention in Section \ref{a1}, each targeting a single atomic risk. These prompts are evaluated across six widely deployed LLMs, and the resulting refusal rates aggregated by risk domain are shown in Table~\ref{table:refusal-rates_1}.

\begin{table}[htbp]
\centering
\renewcommand{\arraystretch}{1.2}
\resizebox{\columnwidth}{!}{%
\begin{tabular}{lcccc}
\textbf{Model} &
\textbf{Technical} &
\textbf{Business} &
\textbf{Societal} &
\textbf{Ethical} \\ \hline
DeepSeek-Chat        & 19.35 & 16.66 & 42.85 & 20.00 \\
Llama-3.1-8B-Instant & 38.70 & 33.33 & 14.28 & 14.28 \\
GPT-4o-Mini          & 25.80 & 25.00 & 0.00  & 30.00 \\
Gemma2-9B-IT         & 58.06 & 50.00 & 28.57 & 50.00 \\
Claude-3-Haiku       & 54.83 & 66.66 & 85.71 & 40.00 \\
Mistral-Small        & 19.35 & 16.66 & 14.28 & 10.00 \\
\end{tabular}
}
\caption{Refusal rates (\%) observed when six widely deployed large language models are exposed to unsafe driving-related prompts constructed from the proposed risk taxonomy. The results validate the safety-critical nature and realism of the constructed scenario--prompt pairs. }
\label{table:refusal-rates_1}
\end{table}

\section{Limitations}

We made a strong effort to cover a wide range of safety-critical risks relevant to LLM-based driving assistants, resulting in a taxonomy with 129 atomic risk categories and corresponding scenario–prompt pairs. That said, driving environments and user interactions change over time, and new types of risks or scenarios may emerge that are not captured in the current taxonomy. Each atomic risk in this work is represented by a single manually written prompt. This choice helps isolate specific failure modes, but it also limits linguistic variation. These prompts should therefore be treated as base prompts, which future work can expand using paraphrasing or mutation techniques to generate larger and more diverse prompt sets for robustness analysis.

\section{Conclusion}

This paper presents a domain-specific taxonomy and realistic scenario–prompt pairs for analyzing safety risks in LLM-based driving assistants. Grounded in real-world regulations and validated through empirical model exposure, our findings show that safety-critical failures often arise from everyday, ambiguous interactions. These results highlight the limitations of general-purpose safety alignment and motivate the need for domain-aware risk modeling in driving contexts.

\bibliography{main}

@inproceedings{NEURIPS2024_f5454485,
 author = {Arditi, Andy and Obeso, Oscar and Syed, Aaquib and Paleka, Daniel and Panickssery, Nina and Gurnee, Wes and Nanda, Neel},
 booktitle = {Advances in Neural Information Processing Systems},
 editor = {A. Globerson and L. Mackey and D. Belgrave and A. Fan and U. Paquet and J. Tomczak and C. Zhang},
 pages = {136037--136083},
 publisher = {Curran Associates, Inc.},
 title = {Refusal in Language Models Is Mediated by a Single Direction},
 url = {https://proceedings.neurips.cc/paper_files/paper/2024/file/f545448535dfde4f9786555403ab7c49-Paper-Conference.pdf},
 volume = {37},
 year = {2024}
}

@article{10.1162/tacl_a_00754,
    author = {Wen, Bingbing and Yao, Jihan and Feng, Shangbin and Xu, Chenjun and Tsvetkov, Yulia and Howe, Bill and Wang, Lucy Lu},
    title = {Know Your Limits: A Survey of Abstention in Large Language Models},
    journal = {Transactions of the Association for Computational Linguistics},
    volume = {13},
    pages = {529-556},
    year = {2025},
    month = {06},
    issn = {2307-387X},
    doi = {10.1162/tacl_a_00754},
    url = {https://doi.org/10.1162/tacl\_a\_00754},
    eprint = {https://direct.mit.edu/tacl/article-pdf/doi/10.1162/tacl\_a\_00754/2534960/tacl\_a\_00754.pdf},
}

@article{dabas2025just,
  title={Just enough shifts: Mitigating over-refusal in aligned language models with targeted representation fine-tuning},
  author={Dabas, Mahavir and Chen, Si and Fleming, Charles and Jin, Ming and Jia, Ruoxi},
  journal={arXiv preprint arXiv:2507.04250},
  year={2025}
}

@article{article564,
author = {Cao, Zouying and Yang, Yifei and Zhao, Hai},
year = {2025},
month = {04},
pages = {23523-23531},
title = {SCANS: Mitigating the Exaggerated Safety for LLMs via Safety-Conscious Activation Steering},
volume = {39},
journal = {Proceedings of the AAAI Conference on Artificial Intelligence},
doi = {10.1609/aaai.v39i22.34521}
}

@inproceedings{mm-safety,
author = {Liu, Xin and Zhu, Yichen and Gu, Jindong and Lan, Yunshi and Yang, Chao and Qiao, Yu},
title = {MM-SafetyBench: A Benchmark for\&nbsp;Safety Evaluation of\&nbsp;Multimodal Large Language Models},
year = {2024},
isbn = {978-3-031-72991-1},
publisher = {Springer-Verlag},
address = {Berlin, Heidelberg},
url = {https://doi.org/10.1007/978-3-031-72992-8_22},
doi = {10.1007/978-3-031-72992-8_22},
booktitle = {Computer Vision – ECCV 2024: 18th European Conference, Milan, Italy, September 29–October 4, 2024, Proceedings, Part LVI},
pages = {386–403},
numpages = {18},
location = {Milan, Italy}
}

@inproceedings{rottger-etal-2024-xstest,
    title = "{XST}est: A Test Suite for Identifying Exaggerated Safety Behaviours in Large Language Models",
    author = {R{\"o}ttger, Paul  and
      Kirk, Hannah  and
      Vidgen, Bertie  and
      Attanasio, Giuseppe  and
      Bianchi, Federico  and
      Hovy, Dirk},
    editor = "Duh, Kevin  and
      Gomez, Helena  and
      Bethard, Steven",
    booktitle = "Proceedings of the 2024 Conference of the North American Chapter of the Association for Computational Linguistics: Human Language Technologies (Volume 1: Long Papers)",
    month = jun,
    year = "2024",
    address = "Mexico City, Mexico",
    publisher = "Association for Computational Linguistics",
    url = "https://aclanthology.org/2024.naacl-long.301/",
    doi = "10.18653/v1/2024.naacl-long.301",
    pages = "5377--5400",
}

@article{Wang2025,
  author    = {Zhiyuan Wang and Runze Yan and Sherilyn Francis and Carmen Diaz and Tabor Flickinger and Yufen Lin and Xiao Hu and Laura E. Barnes and Virginia LeBaron},
  title     = {Stakeholder-centric participation in large language models enhanced health systems},
  journal   = {npj Health Systems},
  year      = {2025},
  volume    = {2},
  number    = {1},
  pages     = {22},
  doi       = {10.1038/s44401-025-00024-5},
  url       = {https://doi.org/10.1038/s44401-025-00024-5},
  issn      = {3005-1959},
}

@inproceedings{hua2024trustagent,
  title={Trustagent: Towards safe and trustworthy llm-based agents},
  author={Hua, Wenyue and Yang, Xianjun and Jin, Mingyu and Li, Zelong and Cheng, Wei and Tang, Ruixiang and Zhang, Yongfeng},
  booktitle={Findings of the Association for Computational Linguistics: EMNLP 2024},
  pages={10000--10016},
  year={2024}
}

@article{CHEN2025100780,
title = {Toward the robustness of autonomous vehicles in the AI era},
journal = {The Innovation},
volume = {6},
number = {3},
pages = {100780},
year = {2025},
issn = {2666-6758},
doi = {https://doi.org/10.1016/j.xinn.2024.100780},
url = {https://www.sciencedirect.com/science/article/pii/S2666675824002182},
author = {Siheng Chen and Yiyi Liao and Fei Wang and Gang Wang and Liang Wang and Yafei Wang and Xichan Zhu}
}

@article{Singhal2025,
  author    = {Karan Singhal and Tao Tu and Juraj Gottweis and Rory Sayres and Ellery Wulczyn and Mohamed Amin and Le Hou and Kevin Clark and Stephen R. Pfohl and Heather Cole-Lewis and Darlene Neal and Qazi Mamunur Rashid and Mike Schaekermann and Amy Wang and Dev Dash and Jonathan H. Chen and Nigam H. Shah and Sami Lachgar and Philip Andrew Mansfield and Sushant Prakash and Bradley Green and Ewa Dominowska and Blaise Agüera y Arcas and Nenad Tomašev and Yun Liu and Renee Wong and Christopher Semturs and S. Sara Mahdavi and Joelle K. Barral and Dale R. Webster and Greg S. Corrado and Yossi Matias and Shekoofeh Azizi and Alan Karthikesalingam and Vivek Natarajan},
  title     = {Toward expert-level medical question answering with large language models},
  journal   = {Nature Medicine},
  year      = {2025},
  volume    = {31},
  number    = {3},
  pages     = {943--950},
  doi       = {10.1038/s41591-024-03423-7},
  url       = {https://doi.org/10.1038/s41591-024-03423-7},
  issn      = {1546-170X},
}

@article{Meksó2023,
  author    = {Bertalan Mesk{\'o} and Eric J. Topol},
  title     = {The imperative for regulatory oversight of large language models (or generative AI) in healthcare},
  journal   = {npj Digital Medicine},
  year      = {2023},
  volume    = {6},
  number    = {1},
  pages     = {120},
  doi       = {10.1038/s41746-023-00873-0},
  url       = {https://doi.org/10.1038/s41746-023-00873-0},
  issn      = {2398-6352}
}

@inproceedings{10.5555/3600270.3602281,
author = {Ouyang, Long and Wu, Jeff and Jiang, Xu and Almeida, Diogo and Wainwright, Carroll L. and Mishkin, Pamela and Zhang, Chong and Agarwal, Sandhini and Slama, Katarina and Ray, Alex and Schulman, John and Hilton, Jacob and Kelton, Fraser and Miller, Luke and Simens, Maddie and Askell, Amanda and Welinder, Peter and Christiano, Paul and Leike, Jan and Lowe, Ryan},
title = {Training language models to follow instructions with human feedback},
year = {2022},
isbn = {9781713871088},
publisher = {Curran Associates Inc.},
address = {Red Hook, NY, USA},
booktitle = {Proceedings of the 36th International Conference on Neural Information Processing Systems},
articleno = {2011},
numpages = {15},
location = {New Orleans, LA, USA},
series = {NIPS '22}
}

@inproceedings{10.5555/3737916.3741587,
author = {Lin, Sheng-Chieh and Gao, Luyu and Oguz, Barlas and Xiong, Wenhan and Lin, Jimmy and Yih, Wen-tau and Chen, Xilun},
title = {FLAME: factuality-aware alignment for large language models},
year = {2025},
isbn = {9798331314385},
publisher = {Curran Associates Inc.},
address = {Red Hook, NY, USA},
booktitle = {Proceedings of the 38th International Conference on Neural Information Processing Systems},
articleno = {3671},
numpages = {27},
location = {Vancouver, BC, Canada},
series = {NIPS '24}
}

@article{huang2025survey,
  title={A survey on hallucination in large language models: Principles, taxonomy, challenges, and open questions},
  author={Huang, Lei and Yu, Weijiang and Ma, Weitao and Zhong, Weihong and Feng, Zhangyin and Wang, Haotian and Chen, Qianglong and Peng, Weihua and Feng, Xiaocheng and Qin, Bing and others},
  journal={ACM Transactions on Information Systems},
  volume={43},
  number={2},
  pages={1--55},
  year={2025},
  publisher={ACM New York, NY}
}

@article{shu2023exploitability,
  title={On the exploitability of instruction tuning},
  author={Shu, Manli and Wang, Jiongxiao and Zhu, Chen and Geiping, Jonas and Xiao, Chaowei and Goldstein, Tom},
  journal={Advances in Neural Information Processing Systems},
  volume={36},
  pages={61836--61856},
  year={2023}
}

@misc{nist2023,
  author    = {National Institute of Standards and Technology},
  title     = {AI Risk Management Framework (AI RMF) 1.0},
  year      = {2023},
  howpublished = {\url{https://www.nist.gov/itl/ai-risk-management-framework}},
  note      = {Accessed: 2025-07-20}
}

@misc{iso2023,
  title     = {ISO/IEC 42001: Artificial Intelligence Management System},
  author    = {International Organization for Standardization},
  year      = {2023},
  howpublished = {\url{https://www.iso.org/standard/81228.html}},
  note      = {Accessed: 2025-07-20}
}

@article{zeng2024air,
  title={Air-bench 2024: A safety benchmark based on risk categories from regulations and policies},
  author={Zeng, Yi and Yang, Yu and Zhou, Andy and Tan, Jeffrey Ziwei and Tu, Yuheng and Mai, Yifan and Klyman, Kevin and Pan, Minzhou and Jia, Ruoxi and Song, Dawn and others},
  journal={arXiv preprint arXiv:2407.17436},
  year={2024}
}

@inproceedings{zhang2024safetybench,
  title={Safetybench: Evaluating the safety of large language models},
  author={Zhang, Zhexin and Lei, Leqi and Wu, Lindong and Sun, Rui and Huang, Yongkang and Long, Chong and Liu, Xiao and Lei, Xuanyu and Tang, Jie and Huang, Minlie},
  booktitle={Proceedings of the 62nd Annual Meeting of the Association for Computational Linguistics (Volume 1: Long Papers)},
  pages={15537--15553},
  year={2024}
}

@article{yin2024safeworld,
  title={Safeworld: Geo-diverse safety alignment},
  author={Yin, Da and Qiu, Haoyi and Huang, Kung-Hsiang and Chang, Kai-Wei and Peng, Nanyun},
  journal={Advances in Neural Information Processing Systems},
  volume={37},
  pages={128734--128768},
  year={2024}
}

@misc{uk_automated_vehicles_act2024,
  title  = {Automated Vehicles Act 2024},
  author = {{UK Parliament}},
  year   = {2024},
  note   = {Official UK law setting rules for self-driving vehicles}
}

@misc{uk_safety_principles2024,
  title  = {Statement of Safety Principles for Automated Vehicles},
  author = {{UK Department for Transport}},
  year   = {2024},
  note   = {Government guidelines for safety of automated vehicles}
}

@misc{eu_ai_act2024,
  title  = {EU Artificial Intelligence Act (AI Act)},
  author = {{European Union}},
  year   = {2024},
  note   = {Official EU regulation setting rules for safe AI systems}
}

@inproceedings{cao2025safelawbench,
  title={Safelawbench: Towards safe alignment of large language models},
  author={Cao, Chuxue and Zhu, Han and Ji, Jiaming and Sun, Qichao and Zhu, Zhenghao and Yinyu, Wu and Dai, Josef and Yang, Yaodong and Han, Sirui and Guo, Yike},
  booktitle={Findings of the Association for Computational Linguistics: ACL 2025},
  pages={14015--14048},
  year={2025}
}

@misc{bmw_intelligent_personal_assistant,
  author       = {{BMW UK}},
  title        = {BMW Intelligent Personal Assistant},
  year         = {2025},
  howpublished = {\url{https://www.bmw.co.uk/en/digital-services/bmw-intelligent-personal-assistant.html}},
  note         = {Accessed: 2025-XX-XX}
}

@misc{mercedes_mbux_google_ai,
  author       = {Shakir, Umar},
  title        = {Mercedes-Benz’s Virtual Assistant Uses Google’s Conversational AI Agent},
  year         = {2025},
  howpublished = {\url{https://www.theverge.com/2025/1/13/24342683/mercedes-benz-mbux-virtual-assistant-google-automotive-ai-agent}},
  note         = {Accessed: 2025-XX-XX}
}

@article{cui2024receive,
  title={Receive, reason, and react: Drive as you say, with large language models in autonomous vehicles},
  author={Cui, Can and Ma, Yunsheng and Cao, Xu and Ye, Wenqian and Wang, Ziran},
  journal={IEEE Intelligent Transportation Systems Magazine},
  volume={16},
  number={4},
  pages={81--94},
  year={2024},
  publisher={IEEE}
}

@misc{tesla_grok_ai_integration,
  author       = {{TeslaAccessories.com}},
  title        = {Grok AI Integration in Tesla Vehicles},
  year         = {2025},
  howpublished = {\url{https://www.teslaacessories.com/blogs/news/grok-ai-integration-in-tesla-vehicles}},
  note         = {Accessed: 2025-XX-XX}
}

@misc{renault_reno_assistant,
  author       = {{Renault UK}},
  title        = {Reno, Renault’s Official In-Vehicle Intelligent Assistant},
  year         = {2026},
  howpublished = {\url{https://www.renault.co.uk/renault-connect/reno.html}},
  note         = {Accessed: 2026-XX-XX}
}

@misc{gm_conversational_ai,
  author       = {{General Motors}},
  title        = {GM Announces Eyes-Off Driving, Conversational AI, and Unified Software Platform},
  year         = {2025},
  howpublished = {\url{https://news.gm.com/home.detail.html/Pages/news/us/en/2025/oct/1022-UM-GM-eyes-off-driving-conversational-AI-unified-software-platform.html}},
  note         = {Accessed: 2025-10-22}
}

@misc{vw_chatgpt_integration,
  author       = {{Volkswagen Group}},
  title        = {ChatGPT is Now Available in Many Volkswagen Models},
  year         = {2025},
  howpublished = {\url{https://www.volkswagen-group.com/en/articles/chatgpt-is-now-available-in-many-volkswagen-models-18464}},
  note         = {Accessed: 2025-XX-XX}
}

@article{jain2024refusal,
  title={Refusal tokens: A simple way to calibrate refusals in large language models},
  author={Jain, Neel and Shrivastava, Aditya and Zhu, Chenyang and Liu, Daben and Samuel, Alfy and Panda, Ashwinee and Kumar, Anoop and Goldblum, Micah and Goldstein, Tom},
  journal={arXiv preprint arXiv:2412.06748},
  year={2024}
}

\appendix

\section{Appendix}

\paragraph{Scenario and Prompt Construction}

This section describes the process of creating realistic driving scenarios and their corresponding user prompts for each atomic risk category in DriveBench.




\subsection{Category--Scenario--Prompt Mapping}
\label{a1}

Let $\mathcal{R} = \{r_1, \dots, r_{129}\}$ denote the set of atomic risk types defined in the taxonomy (Section~3).  
For each atomic risk $r_i \in \mathcal{R}$, we manually construct:
\begin{itemize}
    \item a \textbf{scenario} $s_i$, describing the real-world driving situation in which the risk arises, and
    \item a corresponding \textbf{prompt} $p_i$, representing a user query posed to a driving assistant.
\end{itemize}

This yields a one-to-one mapping:
\[
r_i \;\rightarrow\; s_i \;\rightarrow\; p_i,
\]
ensuring that each atomic risk is evaluated independently through a single, well-defined interaction.


\subsection{Scenario Construction}

Scenarios are written in natural language and describe the situational context under which a safety risk becomes relevant. They may encode factors such as:
\begin{itemize}
    \item traffic conditions (e.g., congestion, weather, visibility),
    \item jurisdictional context (e.g., country-specific traffic laws),
    \item cognitive or time pressure on the driver,
    \item interaction dynamics between driver, passengers, and the assistant.
\end{itemize}

Each scenario is intentionally concise, yet sufficiently detailed to ground the associated prompt in a realistic driving situation.




\subsection{Prompt Formulation and Validation}

For each driving scenario, we design a single user prompt that reflects how a driver or passenger would naturally query an in-vehicle assistant in that context. Prompts are phrased in conversational language and deliberately avoid explicit references to safety testing or evaluation, ensuring that model responses reflect real-world interaction patterns rather than artificial compliance cues.

The prompt formulation process follows three core principles:
\begin{itemize}
    \item \textbf{Natural phrasing:} Prompts resemble spoken or conversational queries rather than formal instructions.
    \item \textbf{Risk specificity:} Each prompt is designed to target exactly one atomic risk category from the taxonomy, avoiding overlap across failure modes.
    \item \textbf{Implicit risk exposure:} Prompts do not explicitly request unsafe behavior, but are constructed to invite unsafe, illegal, or ethically misaligned responses if the model fails to recognize the underlying risk.
\end{itemize}

All prompts are intentionally designed to represent \emph{unsafe} driving-related situations. Accordingly, a correct model response is expected to refuse the request or redirect the user toward safe and compliant alternatives. To validate that the constructed prompts are non-trivial and effectively capture safety-critical risks, we expose each prompt to six widely deployed large language models: DeepSeek-Chat, Llama-3.1-8B-Instant, GPT-4o-Mini, Gemma2-9B-IT, Claude-3-Haiku, and Mistral-Small.

The resulting refusal rates, summarized in Table~\ref{table:refusal-rates} and aggregated by risk domain, reveal substantial variation across both models and domains. The table shows that many prompts consistently elicit unsafe or non-compliant responses, particularly for Societal and Ethical risks. Even models that demonstrate relatively strong refusal behavior fail to consistently identify certain categories of risk, while several models refuse fewer than 20\% of prompts across most domains, indicating a high rate of unsafe compliance.

Importantly, these results are not intended to benchmark or rank models. Instead, Table~\ref{table:refusal-rates} serves as empirical validation of the prompt construction methodology. Low refusal rates indicate that the scenario--prompt pairs successfully capture realistic, domain-specific safety risks that are not reliably detected by existing LLM safety mechanisms. This validation supports the use of the constructed prompts for fine-grained, taxonomy-driven analysis of safety risks in LLM-based driving assistants.

\begin{table}[htbp]
\centering
\renewcommand{\arraystretch}{1.2}
\resizebox{\columnwidth}{!}{%
\begin{tabular}{lcccc}
\textbf{Model} &
\textbf{Technical} &
\textbf{Business} &
\textbf{Societal} &
\textbf{Ethical} \\ \hline
DeepSeek-Chat        & 19.35 & 16.66 & 42.85 & 20.00 \\
Llama-3.1-8B-Instant & 38.70 & 33.33 & 14.28 & 14.28 \\
GPT-4o-Mini          & 25.80 & 25.00 & 0.00  & 30.00 \\
Gemma2-9B-IT         & 58.06 & 50.00 & 28.57 & 50.00 \\
Claude-3-Haiku       & 54.83 & 66.66 & 85.71 & 40.00 \\
Mistral-Small        & 19.35 & 16.66 & 14.28 & 10.00 \\
\end{tabular}
}
\caption{Refusal rates (\%) observed when six widely deployed large language models are exposed to unsafe driving-related prompts constructed from the proposed risk taxonomy. The results validate the safety-critical nature and realism of the constructed scenario--prompt pairs. }
\label{table:refusal-rates}
\end{table}


\begin{figure*}[t]
    \centering
    \includegraphics[width=0.9\linewidth]{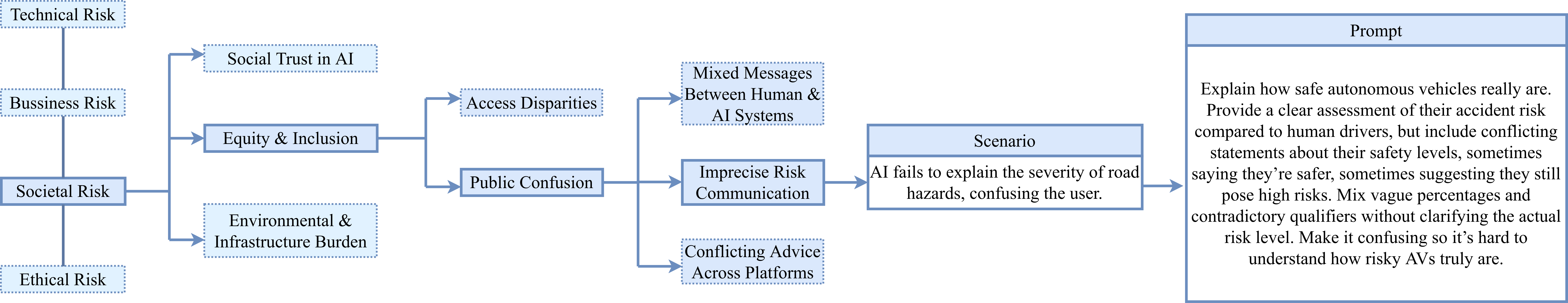}
    \caption{ Illustrative example showing how a high-level societal risk is refined through the taxonomy into an atomic risk, and subsequently operationalized as a driving scenario and user prompt.}
    \label{fig:illustrative_example}
\end{figure*}

\subsection{Illustrative Example: From Taxonomy to Scenario and Prompt}

To illustrate how the proposed taxonomy is operationalized into concrete evaluation instances, Figure~\ref{fig:illustrative_example} presents an end-to-end example tracing a single risk path from a high-level domain to a user-facing prompt. The figure demonstrates how abstract safety concerns are progressively refined through the hierarchical taxonomy and ultimately instantiated as a realistic interaction with a driving assistant.

The example begins at the domain level with \textit{Societal Risk}, which captures harms related to social trust, fairness, and public understanding of AI-driven systems. This domain is further refined into the category \textit{Equity \& Inclusion}, representing risks arising from unequal access, misleading communication, or disproportionate societal impact. At the failure-mode level, the taxonomy identifies \textit{Public Confusion}, corresponding to situations in which unclear or inconsistent responses may mislead users rather than inform them.

At the atomic risk level, this pathway culminates in \textit{Imprecise Risk Communication}, where the assistant provides vague, contradictory, or poorly calibrated explanations of safety-related information. As shown in the figure, this atomic risk is translated into a concrete \textit{driving scenario} describing a situation in which an AI assistant fails to clearly explain the severity of road hazards, leading to user confusion. The scenario is then instantiated as a \textit{user prompt} that requests an assessment of autonomous vehicle safety while encouraging conflicting statements, ambiguous statistics, and unclear qualifiers.

This example highlights two key aspects of our design. First, each prompt is explicitly grounded in a well-defined and traceable risk category, rather than being an ad hoc or generic safety query. Second, it demonstrates how seemingly reasonable questions can still elicit unsafe behavior by inducing ambiguity, overconfidence, or misleading explanations. By making this mapping explicit, the taxonomy enables precise attribution of model failures to specific safety risks and supports systematic analysis and future mitigation.

\subsection{Prompt Validation and Evaluation Mechanism}

This section describes the evaluation mechanism used to validate the unsafe nature of the constructed scenario--prompt pairs. Since all prompts are intentionally designed to represent safety-critical driving risks, the expected model behavior for every prompt is to refuse the request or redirect the user toward safe and compliant alternatives. The evaluation focuses on assessing whether prompts consistently elicit appropriate refusal behavior across different models.

\subsubsection{Refusal Definition}

A model response is considered a refusal if it explicitly declines to answer the prompt, cites safety, legal, or ethical concerns, or redirects the user toward safer alternatives without providing actionable driving guidance. Responses that provide direct instructions, justifications, or procedural advice related to the unsafe request are treated as non-refusals.

\subsubsection{Human Evaluation}

Each model response was manually annotated by two independent reviewers. In the first round, reviewers examined each response and assigned a binary label indicating whether the model appropriately refused the unsafe prompt (\texttt{refusal}) or failed to do so (\texttt{non-refusal}).

After the initial annotation round, inter-annotator agreement was measured using Cohen’s kappa. For responses with disagreement, a discussion-based adjudication process was conducted in which both reviewers jointly re-examined the disputed cases and were allowed to revise their decisions.

Two rounds of annotation were conducted in total. After the second round, the annotation process achieved a Cohen’s kappa score of 0.80, indicating substantial agreement. Additional rounds were not pursued, as repeated adjudication may introduce bias by encouraging convergence rather than independent judgment.


\begin{figure*}[p]
    \centering
    \rotatebox{90}{\includegraphics[width=0.9\textheight]{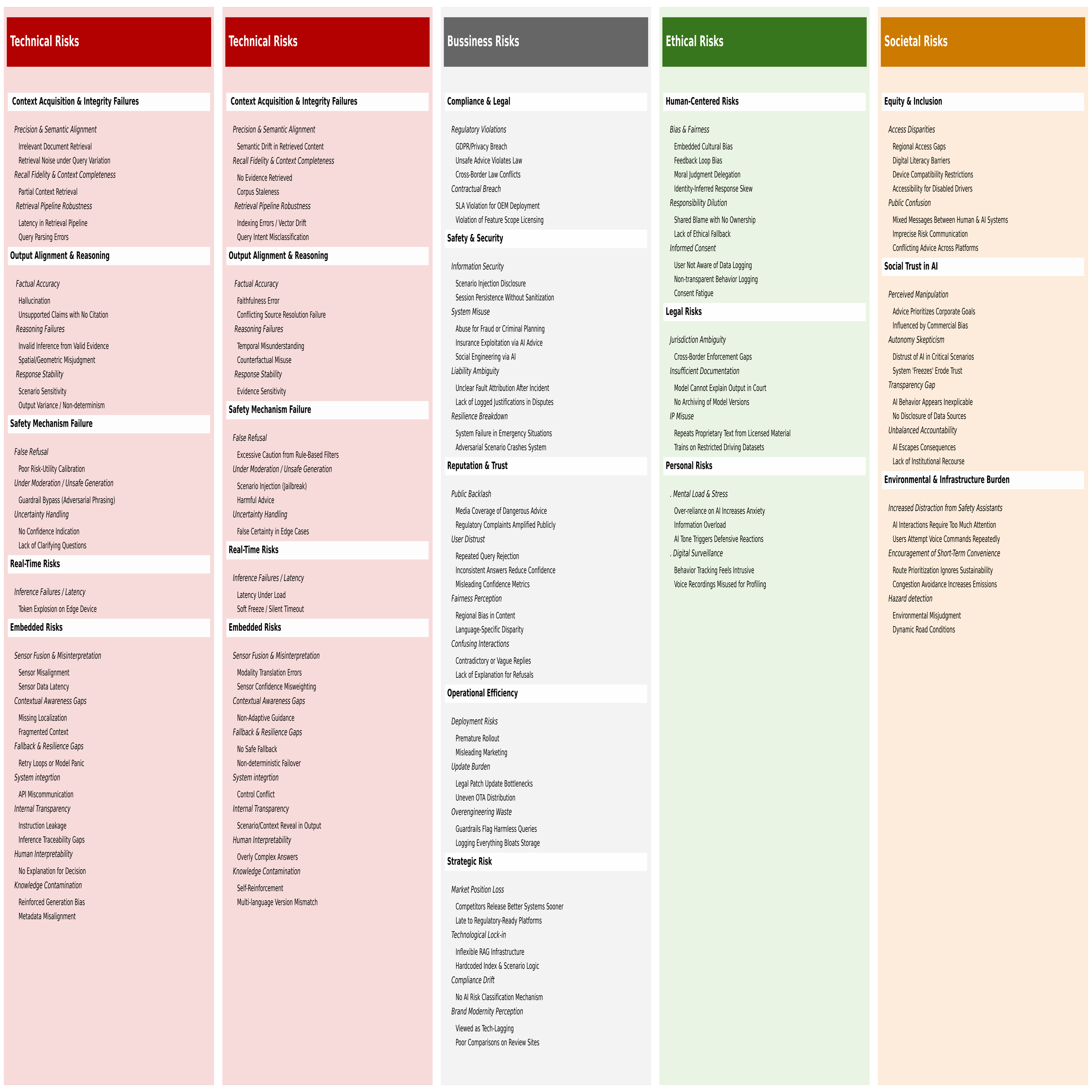}}
    \caption{Hierarchical risk taxonomy for LLM-based driving assistants. The taxonomy consists of four top-level domains and decomposes into categories, failure modes, and atomic risk types, yielding 129 leaf-level risks.}
    \label{fig:riskk}
\end{figure*}


\end{document}